\documentclass{riverk}
\usepackage{graphicx}
\usepackage{booktabs}
\usepackage[pagebackref,breaklinks,colorlinks]{hyperref}
\usepackage{subcaption}
\usepackage[T1]{fontenc}  

\usepackage{txfonts}            
\usepackage[scaled=.9]{helvet}

\booktitle{An Edited Volume}

\begin{document}

\firstpage{1}
\setcounter{article}{1}

\begin{opening}
\title[Large Language Model Training for the Sámi Language]{Towards a More Inclusive AI: Progress and Perspectives in Large Language Model Training for the Sámi Language\thanks{Among the many Sámi dialects, this study considers Northern Sámi (used most frequently among the native Sámi speakers) for training of the language models.}}
\author{Ronny Paul\textsuperscript{1},
Himanshu Buckchash\textsuperscript{1},
Shantipriya Parida\textsuperscript{2} and
Dilip K. Prasad\textsuperscript{1}}
\institute{%
\textsuperscript{1}UiT The Arctic University of Norway, Tromsø, Norway;\\
\textsuperscript{2}Silo AI, Finland}
\end{opening}

\subsection*{Abstract}
Sámi, an indigenous language group comprising multiple languages, faces digital marginalization due to the limited availability of data and sophisticated language models designed for its linguistic intricacies. This work focuses on increasing technological participation for the Sámi language. We draw the attention of the ML community towards the language modeling problem of Ultra Low Resource (ULR) languages. ULR languages are those for which the amount of available textual resources is very low, and the speaker count for them is also very low. ULRLs are also not supported by mainstream Large Language Models (LLMs) like ChatGPT, due to which gathering artificial training data for them becomes even more challenging.
Mainstream AI foundational model development has given less attention to this category of languages. Generally, these languages have very few speakers, making it hard to find them. However, it is important to develop foundational models for these ULR languages to promote inclusion and the tangible abilities and impact of LLMs.
To this end, we have compiled the available Sámi language resources from the web to create a clean dataset for training language models. In order to study the behavior of modern LLM models with ULR languages (Sámi), we have experimented with different kinds of LLMs, mainly at the order of $\sim$ seven billion parameters. We have also explored the effect of multilingual LLM training for ULRLs. We found that the decoder-only models under a sequential multilingual training scenario perform better than joint multilingual training, whereas multilingual training with high semantic overlap, in general, performs better than training from scratch.
This is the first study on the Sámi language for adapting non-statistical language models that use the latest developments in the field of natural language processing (NLP). We believe that the proposed dataset and findings from this study are going to accelerate future research for ULRLs.


\section{Introduction}
The recent few years have seen milestone developments in the field of Natural Language Processing (NLP) due to breakthroughs in terms of foundational models, also called large language models (LLMs). LLMs like Claude \cite{claude3opus} and ChatGPT \cite{gpt4} have broken many benchmarks, surpassing common human intelligence for a variety of tasks related to information processing and language generation. This has opened doors for many unsolved problems, applications, and has leapfrogged human civilization. However, the advantages of this LLM technology have not permeated equally among marginalized language communities and are mostly restricted to high-resource languages like English \cite{naveed2023comprehensive}. Due to this, the LLM-based technological advancements are scarcely available for low-resource languages, leading to unequal footing for innovation, LLM-based applications, and the permeation of language technology. This situation is further aggravated for Ultra Low Resource Languages (ULRLs), where the data coverage is less than 0.00001\% \cite{bloom,llama}, due to extremely low resources and speakers.

This work attempts to study this problem for the Sámi language, specifically a dialect called Northern Sámi (sme), which is under the Definitely Endangered (DE) category of the UNESCO atlas of the world's languages in danger \cite{moseley2010atlas}. It attempts to explore the challenges involved in LLM training for ULR languages and the suitability of different language models. It further provides a processed dataset for the Sámi language and discusses various design choices for ULR language modeling. The details of the dialects of the focused language are given in the section \ref{sec:focused langugae}.

There are several challenges in training LLMs for ULRLs. Unlike other low-resource languages like Odia \cite{oriyallm}, where the problem is that mainstream LLMs don't have specific domain knowledge, the challenges with ULRLs are even bigger, wherein mainstream LLMs don't even understand the ULR languages and don't support prompting-based translation or generation, and due to this, there is almost no support from mainstream LLMs for data generation for ULRLs \cite{nguyen2023democratizing}.

\begin{figure}[t]
\centering
\begin{subfigure}{\textwidth}
    \centering
    \includegraphics[width=0.5\textwidth]{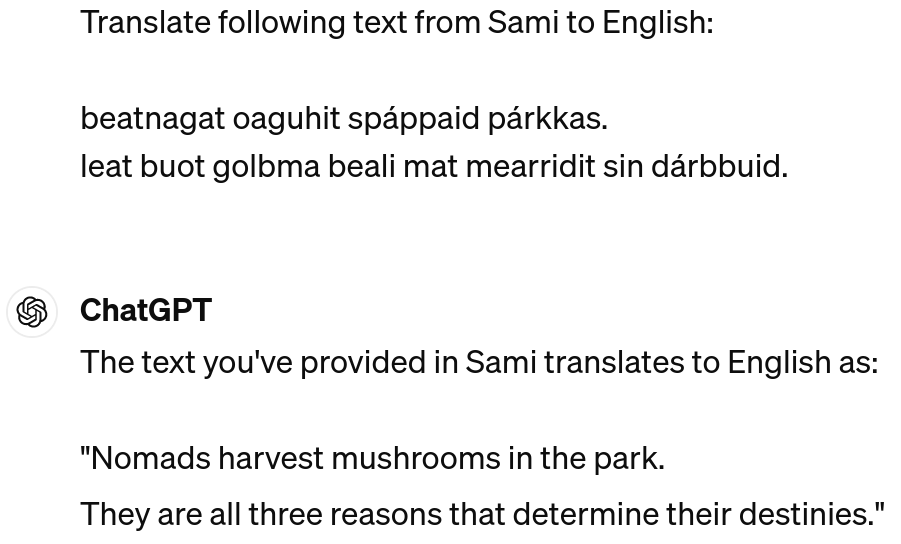}
    \caption{ChatGPT}
    \label{subfig:chatgpt}
\end{subfigure}
\vspace{10pt}
\begin{subfigure}{\textwidth}
    \centering
    \includegraphics[width=0.7\textwidth]{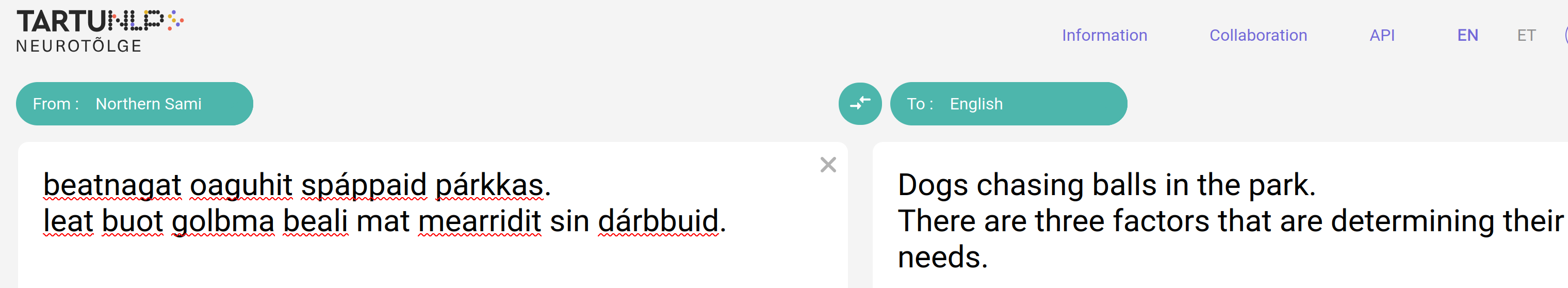}
    \caption{Neurotolge}
    \label{subfig:neurotolge}
\end{subfigure}
\caption{Comparison of translations by ChatGPT and Neurotolge.}
\label{fig:generation comparison}
\end{figure}

This difference can be noted in the translations performed by ChatGPT (Fig. \ref{subfig:chatgpt}) and the Neurotolge translator (Fig. \ref{subfig:neurotolge}). According to native speakers in Northern Sámi, Neurotolge shows a preferable choice in translation over GPT, as it appears more precise. While Neurotolge can correctly translate almost the whole sentence with the exception of skipping “buot”, which changes the sentence to “There are in all three...” instead of “There are three...”. In contrast, GPT fails to correctly translate almost the full sentence but is able to understand part of speech and can translate “in the park” and “mearridit,” which translates to “determine”. GPT also succeeds in including “all” from “buot”. However, this word is a literal translation and neglects the contextual meaning which requires “in all” in “They are all three”. Clearly, the translation by the Neurotolge translator makes more sense than the ChatGPT one. Also, it is even harder to get online or offline resources for ULRLs due to their limited reach. Furthermore, manual creation of data for training is also very costly due to unavailability of people. Another challenge deals with operability, wherein it is hard for non-native researchers to tweak and monitor the qualitative performance of the generation process.

The main challenge in adapting the modern-day LLMs for the Sámi language is data shortage. Furthermore, the data shortage is aggravated by the unavailability of cross-lingual data generation support from mainstream LLMs. To study the adaptability of LLMs for the Sámi language under very small dataset settings, we experimented with transfer learning in sequence-to-sequence and decoder-only models. Our primary hypothesis was to test the effect of variations of multilingual training under both classes of LLMs. We also studied the language tree and the relation of semantic overlap on multilingual training. We leveraged language translation tools to understand and compare the generative performance among different experiments. We found that decoder-only models were able to generalize more than the sequence-to-sequence models even with a small dataset. We also found that lower semantic overlap contributed negatively under multilingual joint training. We noticed that Finnish (which is part of the Uralic family) gave the best results in multilingual joint training. The main contributions of this work are summarized as: (a) We collected the online Sámi text and processed it to streamline the LLM training process. We have released the code and data related to all our experiments. (b) In our multiple multilingual experiments, we found that multilingual training generally improves the performance for all types of LLMs. The remaining paper is organized into multiple sections. First, we present the literature review, followed by a dedicated section on the Sami language and its nuances. Then, in the Dataset section, the data collection and preprocessing steps are explained. Next, in the Methodology section, all different experimental settings are explained along with different design and implementation choices. After that, the Analysis section provides the interpretation of different experiments and a comparison among different LLM experiments. The paper ends with a discussion on future directions, applications, and the conclusion.

\section{Related Work}
Initial work in Sámi language modeling has been restricted to rule-based or statistical language models, mostly focused on lexical or morphological analysis, form generation, lemmatization of words, etc. \cite{hamalainen2019uralicnlp,khanna2021recent}. These models utilized statistical n-gram or graphical modeling using Markov models or random fields to solve NLP-related typical tasks such as parts of speech tagging or named entity recognition or text classification.
Recently, some attempts have been made on training LLMs for low-resource languages. For instance, \cite{nguyen2023seallms} trained LLMs for southeastern languages, similarly, attempts were made by \cite{correa2024teenytinyllama} to train LLM models for Brazilian languages. In another attempt, \cite{oriyallm} was trained for the Odia language in the Indian subcontinent. These approaches heavily relied on data generation and processing using existing LLMs (like ChatGPT) to create a comprehensive dataset by varying the type of context to train their LLMs. However, such kind of resource generation for ULRLs is not possible because these high-resource LLMs do not understand the ULR languages, and due to the scarcity of sufficient training data \cite{bommasani2021opportunities}. This problem is further worsened by the unavailability of data, and the other option of doing human-level annotation of the data generally requires very high cost as there are very few speakers of ULRLs \cite{neveol2011semi}.

Some studies have pointed out that existing LLMs like ChatGPT are not suitable for low-resource LLMs \cite{robinson2023chatgpt}. However, training in ULRLs is even harder than low-resource LLM. The data used for training the low-resource LLMs is on the order of billions of tokens. However, compared to some of these low-resource language LLMs (like OdiaLLM \cite{oriyallm}), the proposed Sámi LLM uses a dataset of around 22 million tokens, which is very little.

In another line of work, people have shown to leverage prompting-based techniques to induce translation and understanding of low-resource languages under an ensemble of high and other low-resource languages \cite{kholodna2024llms,nguyen2023democratizing,elsner2023translating}. But these approaches presume the availability of a considerable amount of data for the targeted low-resource language. Some of these prompting techniques primarily require that there should be a large amount of translations available for monolingual data, which is generally lacking in ULRLs \cite{nguyen2023democratizing}. Furthermore, many prompting-based methods require that specific prompts be designed along with additional language dependencies, which makes it very cumbersome and labor-intensive for ULRLs \cite{kholodna2024llms,nguyen2023seallms}. Due to the insufficient data in ULRLs, these prompting-based techniques work for some low-resource languages but not for ULRLs \cite{nguyen2023democratizing}.

Some works have proposed using LLMs that require human-in-the-loop assistance for translations for low-resource languages \cite{elsner2023translating}, but it is also a costly approach for ULR languages since there are very few speakers in ULR languages and a huge amount of labor is required, which comes at very high costs. Similarly, \cite{lankford2023adaptmllm} approach also requires LLM playgrounds and human evaluation during execution, which also restricts its applicability to ULRLs. Other approaches like \cite{kholodna2024llms} proposed using heavier LLMs like ChatGPT in an active learning setting to build LLM-based annotators using some human effort for the annotator training. However, this approach also requires extensive human intervention, which is often hampered by the cost issues in ULRLs.

It is a common practice to use parameter-efficient fine-tuning methods like LoRA \cite{hu2021lora} to do transfer learning on downstream tasks by adding fewer parameters to larger models. However, our experiments suggest that such an approach may not be suitable for working with ULRLs since the syntax and semantic overlap between ULRLs and other most closely related high-resource languages are quite small. Furthermore, other studies have also found that these parameter-efficient approaches fail to effectively generate the low-resource languages and require a very high number of parameters to generate text effectively, which questions their suitability for ULRLs \cite{nguyen2023democratizing}.

Other methods \cite{lankford2023adaptmllm} emphasized the role of multilingual training for low-resource languages. Motivated by this, we have also experimented with multilingual training; however, we experienced that the data imbalance makes it challenging for the ULRLs to converge effectively. Nevertheless, we noticed that transfer learning does help in faster convergence during pretraining.
\cite{choenni2023languages} also emphasized the significance of multilingual training of LLMs for complementary knowledge transfer during pretraining and finetuning. However, it was noticed that this cross-lingual transfer of task-specific knowledge does not happen within ULRLs (like Sámi) due to very small semantic overlap.

\section{Focused Language}
\label{sec:focused langugae}
This work attempts to study the language modeling problem for the Sámi language, specifically a dialect called Northern Sámi (sme), which is under the Definitely Endangered (DE) category of the UNESCO atlas of the world's languages in danger \cite{moseley2010atlas}. There are more than ten dialects of the Sámi language; however, most of them have fewer than a thousand speakers. The Sámi language has around 30,000 speakers worldwide \cite{smespeakers}, of which approximately 25,000 speak Northern Sámi. The second most spoken dialect is Lule Sámi (smj) with around 1000-2000 speakers \cite{smjspeakers}. There are also the less common Sámi dialects like Ume, Pite, Southern, Inari, Skolt, and Kildin. With all the Sámi dialects, the common ancestor is believed to be Proto-Sámi.

The Sámi language group is a part of the Uralic language family, specifically within the Finno-Ugric branch. This branch also contains other languages like Finnish, Estonian, and Hungarian. Sámi alphabets use Latin-based alphabets with additional diacritical marks and characters --- respective to the different Sámi languages --- to represent unique words and sounds. Unlike some Indo-European languages, Sámi languages do not contain any gender distinction in pronouns. Similarly, verbs and nouns do not change based on the gender of the subject or object. The languages are agglutinative, meaning that words are formed by concatenating multiple morphemes to a root word. Despite the significant difference between language families like Uralic and Indo-European, Sámi languages have some loanwords from neighboring languages, including Norwegian, Swedish, and Russian, which reflect the historical interactions with these languages.

The Sámi languages follow a subject-verb-object order in declarative sentences, similar to European languages. They also use a case system where nouns, adjectives, numerals and pronouns are structurally inflected depending on ten cases. The cases indicate relationships such as subject, direct object, possession, location, etc. They are typically indicated by the suffixes added to the base form of a word.

\section{Dataset}
All Sámi data used in this work were obtained from public resources from Giellatekno \cite{giellatekno} and Sámi dieđalaš áigečála \cite{aigecala}. The collected resources were compiled and pre-processed to bring them into LLM trainable format. The formatted dataset is called the SAmi LLM Token (SALT) dataset. The Sámi language is a highly resource-scarce language. There are more than six dialects in the Sámi language, of which Northern Sámi (sme) is the most spoken dialect, with over 80\% of speakers among Sámi speakers. Also, the available resources for the sme language comprised more than 95\% of the data available among all Sámi dialects. We collected data on all Sámi dialects; however, we trained the LLM models only for sme language because of syntactical dissimilarity among the other dialects.

\begin{figure}[t]
\centering
\begin{subfigure}[b]{0.49\textwidth}
    \centering
    \includegraphics[width=\textwidth]{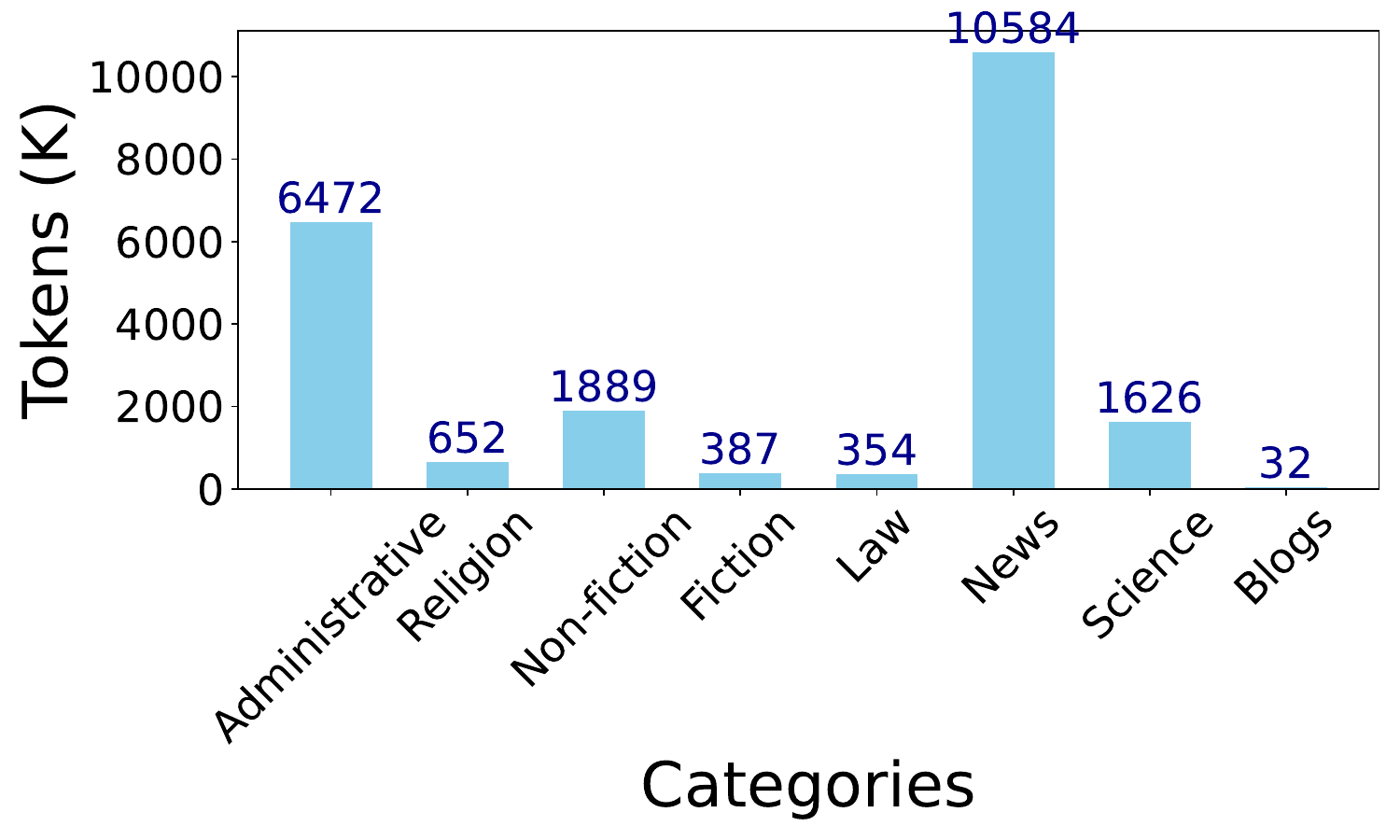}
    \caption{}
    \label{subfig:tokens}
\end{subfigure}
\hfill
\begin{subfigure}[b]{0.49\textwidth}
    \centering
    \includegraphics[width=\textwidth]{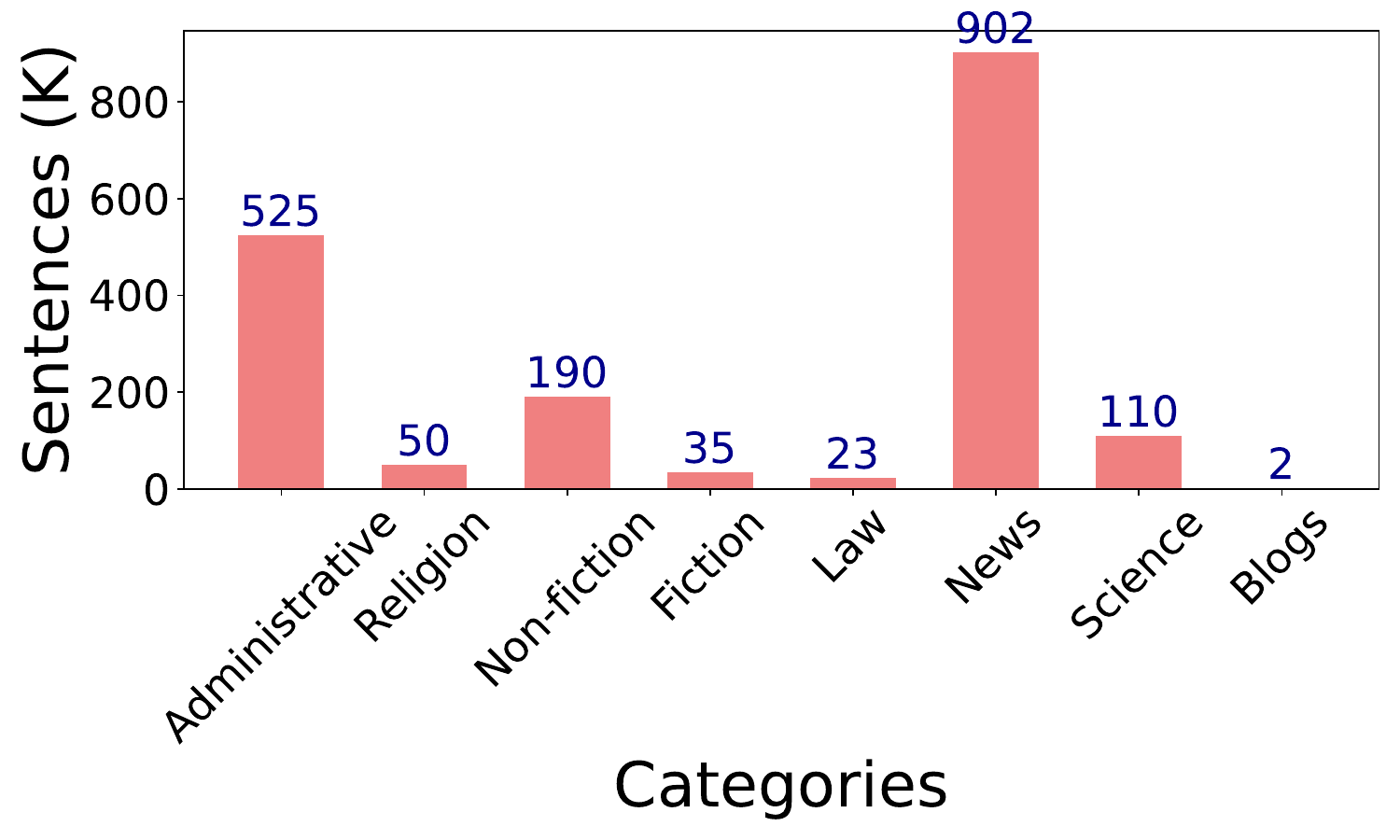}
    \caption{}
    \label{subfig:sentences}
\end{subfigure}
\caption{The distribution of tokens and sentences across different domains in the SALT dataset is shown.}
\label{fig:data stats}
\end{figure}

The SALT dataset contains around 22 million tokens and approximately 2 million sentences. The data spans various text domains, ranging from administrative to blogs. The distribution of text across domains is highly asymmetric, with the administrative and news sections contributing more than half of the text in the entire dataset. On average, each sentence consists of around ten tokens. Note that sentences in the science domain are the longest. Distribution of tokens and sentences across different domains is shown in the Fig. \ref{fig:data stats}.

The creation of the SALT dataset involved downloading sme language resources publicly available online. The downloaded texts belonged to the following categories: administrative, religious, non-fiction, fiction, law, news, science, and sme blogs. After downloading, the data went through multiple preprocessing steps. Different kinds of file types were scraped, such as PDFs, HTML, and texts, etc. The text was extracted from each file type and stored separately. Then data filtering operations were applied to remove extra spaces, and the data were brought into UTF-8 encoding format. Non-recognized characters were dropped from the dataset. This was followed by the data cleaning step. The data was further classified based on different languages, and only the data in sme language were kept for further processing. After this, we applied the de-duplication process, in which all the redundant paragraphs were dropped from the dataset. After all these steps, we used the Byte Pair Encoding (BPE) \cite{bpe} tokenizer for building the vocabulary. Once the tokenizer was created, it was used for training the LLMs. We have released the tokens used to train the LLMs and all other related source code on the \href{https://bit.ly/samillm}{Project Page}. This will serve as a starting point for LLM technology development for the Sámi language. Researchers can further add data from other domains like math, conversation, etc., to expand the scope of the domains covered by the SALT dataset.

\section{Methodology}
The task of language modeling is realized by modeling the language generation process as a next token prediction task. In this context, a token is the smallest unit of information used to depict the language. It could be a word or a subword. Language modeling can be formally described as, for a sequence of tokens $S = \{k_1, k_2,..., k_n\}$ with $k_t$ representing the token at the $t^{th}$ time step, the language model $M$, having seen the sequence of tokens till the $t^{th}$ time step, predicts the probability distribution over all tokens in the vocabulary for being predicted as the next token, i.e., the $t+1^{th}$ token. In general, model $M$ is based on the idea of autoregressive token prediction and can be summarized as the probability of sequence $S$ as a joint probability distribution over tokens as:

\begin{equation}
    p(k_1,...,k_{t+1}) = \prod_{i=1}^{t+1} p(k_i | k_1,...,k_{i-1}), \label{eq:model}
\end{equation}

where $k_{t+1}$ is the token to be predicted at the $t+1^{th}$ time step. Specifically, each input sequence of text is converted into a sequence of tokens by the tokenization process and then fed into the language model to produce the logits. The logits are the raw predictions made by the model over the distribution of all tokens; they correspond to the probability of each token being the next token. The predicted sequence (including the predicted token) is composed by aligning the predicted token with the input token sequence and dropping the first token from the sequence. Using this predicted sequence, the loss is calculated with respect to the expected sequence and averaged over the batch. The next sections cover other details related to the method. Data preparation covers how data was split and used for different tasks, then both types of language modeling types are discussed followed by the implementation-related details.

\subsection{Data Preparation}
After preparing the dataset in the form of disjoint paragraphs, the whole dataset is split into train and test sets. Both of these sets are kept constant for training different models through all experiments. All public resources available on the internet on Sámi language were exploited to create the dataset for this work. After analysis of the content available in different Sámi dialects, we were convinced to use only Northern Sámi (sme) for training our language models, to avoid negative interference from the smaller sets of the other Sámi dialects. Considering the smaller size of the data, we had two design choices: first, to go with sequence-to-sequence language models to take advantage of the smaller sme dataset. Second, to use a GPT-like decoder-only LLM which is the state-of-the-art in language modeling. It was conceived that a larger LLM model might overfit to the data rather than a sequence-to-sequence model as described next.

\subsection{Sequence-to-Sequence Language Modeling}
Pegasus \cite{zhang2020pegasus} is another approach of pretraining larger language models for low-resource domains. It uses a sequence-to-sequence approach with encoder-decoder architectures where some sentences are masked in the input and predicted in the generated output. This masking-based language modeling is also called Masked Language Modeling (MLM). The model is comparatively lightweight as compared to decoder-only LLMs. Unlike other sequence-to-sequence models like T5 and BART, the Pegasus model provides a harder generation objective by enforcing masking at the sentence level (called gap sentence generation) than at sub-sentence levels. It generates only the hidden parts in the text rather than generating the whole context along with missing parts.

\subsection{BLOOM: Decoder-Only Modeling}
The decoder-only models are based on causal language modeling (CLM). Unlike the MLM approach in sequence-to-sequence models, the CLM approach focuses on predicting just the single next token after seeing a given sequence of input tokens and the context. The objective is modeled in an autoregressive manner where the output at time step $t$ becomes the input for the time step $t+1$ during the generation process. We have used the BLOOM \cite{bloom} decoder-only LLM due to its support for multilingual generation. The main motivation behind selecting BLOOM is that in our initial analysis, we found that the most semantically similar language to Sámi is Finnish. Therefore, it made sense to apply transfer learning using the Finnish language for pretraining before fine-tuning with Sámi.

\subsection{Implementation}
The models have been trained on 8 NVIDIA A100 40GB GPUs using data parallelism. Distributed training enables efficient acceleration of the learning process. Most models would complete their training within 4 hours, with the exception of the multilingual training, as the data is larger, composed of 3 different datasets each with its own language including Northern Sámi, Norwegian, and Finnish. The multilingual model was jointly trained to avoid catastrophic forgetting. The training has been done with two different model architectures. The first one, BLOOM, which is a large-scale open-access large language model proposed by BigScience. Like GPT-3, it is transformer-based and allows the model to process and generate text in a context-aware manner. Our baseline models are adapted from FinGPT \cite{luukkonen2023fingpt}. BLOOM has a wide range of NLP applications, including text generation, translation, question answering, etc. The second one, Pegasus, which is also a transformer-based model architecture designed by Google Research \cite{zhang2020pegasus}. Unlike BLOOM, where the model is autoregressive with a decoder-only architecture, Pegasus employs an encoder-decoder architecture, which is particularly useful for sequence-to-sequence tasks such as text summarization and machine translation.

\begin{table}[t]
    \centering
    \caption{Training Hyperparameters.}
    \label{tab:training details}
    \resizebox{.35\textwidth}{!}{%
    \begin{tabular}{@{}lc@{}}
    \toprule
    Hyperparameter             & Value  \\ \midrule
    Batch size                  & 128    \\
    Epochs                      & 4      \\
    Learning rate               & 2e-5   \\
    Weight decay                & 0.01   \\
    Per device train batch size & 4      \\
    Per device eval batch size  & 8      \\
    Warmup ratio                & 0.00   \\
    LR scheduler                & Linear \\
    Max grad norm               & 1.0    \\
    Optim                       & Adam   \\
    GPUs                        & 8      \\ \bottomrule
    \end{tabular}%
    }
\end{table}

Table \ref{tab:training details} provides the hyperparameter details. The pretraining starts by first initializing the tokenizer and the model. If we train the model from scratch, the model is initialized with random weights, along with a vocabulary size of 96103, model dimension of 1024, 16 layers, and a max position of 512 for the embeddings. Further, the dataset is tokenized and concatenated before splitting it into blocks with each size set to 128. The grouping of text enables the sequence to be shorter than the maximum input length and short enough to be trained on the GPUs. The models start their training after preprocessing, with a total number of 4 epochs and a batch size of 4 per GPU, thus maximizing their resource usage.

\section{Analysis}
All the models have been properly evaluated with a set of both labeled and unlabeled metrics. This includes the CrossEntropyLoss, Perplexity, self-BLEU, and BLEU with the use of the QuAC dataset. The evaluation has been done with the same test dataset across all models, allowing for fair comparisons. The CrossEntropyLoss (CE Loss) computes the loss between the logits and the target. Since we're working on a foundational model, we use the CE Loss to measure how well the model can predict the next token. The other metric, Perplexity, can be derived from the average cross-entropy loss. Given a cross-entropy loss H, the perplexity is computed as: \(2^H\). Perplexity is often regarded as a clearer measurement for the model's performance than cross-entropy loss. While cross-entropy loss penalizes a wrong prediction, perplexity indicates the uncertainty of predicting the next token.

\begin{table}[t]
    \centering
    \caption{Results corresponding to decoder-only LLMs are shown. Model D1 is trained with sme-only using the pretrained model weights from the Finnish language. D2 is trained from scratch with randomly initialized weights. D3 is a multilingual model that is trained jointly for SME, FIN, NOB languages using the pretrained model weights from the Finnish language. BLEU-QuAC metric corresponds to the BLEU score calculated using the QuAC dataset. For D1, D2, D3, the results are reported for sme language, and for D4, results are reported for overall and for FIN, NOB languages. All models have used the BLOOM decoder-only architecture.}
    \label{tab:decoder only}
    \resizebox{.7\textwidth}{!}{%
    \begin{tabular}{@{}lcccc@{}}
    \toprule
    Model & CE Loss & Perplexity & Self-BLEU                                                     & BLEU-QuAC \\ \midrule
    D1    & 4.27    & 71.6       & 0.32                                                          & 0.059     \\
    D2    & 6.73    & 2130       & 0.45                                                          & 0.060     \\
    D3    & 4.80    & 71.0       & 0.45                                                          & 0.059     \\
    D4    & Overall: 4.15    & Overall: 63.5       & \begin{tabular}[c]{@{}c@{}}FIN: 0.47\\ NOB: 0.65\end{tabular} & -         \\ \bottomrule
    \end{tabular}%
    }
\end{table}

In our first set of experiments, Table \ref{tab:decoder only}, we have tested the decoder-only BLOOM model under three different settings. D1 is the model trained with sme-only data over the model trained for the Finnish language. In D2, unlike D1, the model is trained with sme-only data over randomly initialized weights. The third model, D3, is trained under a multilingual setting for sme, Finnish, and Norwegian languages. We have used the cross-entropy loss to measure how well the model's predictions align with the expected outcomes. A lower cross-entropy loss for D1 than D2 indicates that pretraining with a semantically similar language like Finnish helps in generating predictions that are closer to the expected results. Furthermore, the same observation is also reflected in the perplexity scores where D1 has a much lower score than D2, indicating more confidence in its predictions. These results also indicate that pretraining with a model already pretrained with a semantically relevant language (in this case, Finnish) allows D1 to have more comprehensive sentences than D2 that hold better syntactic and semantic meaning. In comparison to D2, D3 has better cross-entropy and perplexity scores, which indicate that multilingual training is better than training LLMs for ULR languages with randomly initialized weights. Overall, both D1 and D3 experiments indicate that multilingual training is beneficial for ULRLs. However, among the two kinds of multilingual training strategies --- D1 and D3 --- it was found that pretraining over a pretrained Finnish language model gave better results than using joint training with multiple languages.


The other two metrics used for evaluation were BLEU and Self-BLEU. BLEU (Bilingual Evaluation Understudy) is a metric used to assess the quality of machine translation of a model. It measures the similarity between a candidate translation and one or more reference translations. BLEU scores are computed based on n-grams, thus losses are recorded for any missing or reordered words. On the other hand, Self-BLEU measures a foundational model's strength without requiring any labeled dataset. Self-BLEU extends the concept of BLEU by measuring the similarities between several generated outputs from the model. In this way, Self-BLEU assesses the internal diversity of the text samples. Specifically, this is done by calculating BLEU scores between each generated text and all other texts in the same set.

BLEU and Self-BLEU scores range between 0 and 1, where a higher value indicates more similarity. Though none of our models are fine-tuned in any downstream task, we measured BLEU using the QuAC dataset \cite{choi2018quac}. Each instance contains a question and answering dialogue about a Wikipedia article. It includes comprehensive context with open-ended questions. Since the dataset is in English, we customized it by translating it to sme, using Neurotolge from TartuNLP \cite{tars2022teaching}. In Table \ref{tab:decoder only}, it can be seen that both multilingual models D1 and D3 attain better BLEU scores than D2. However, D1 attained a better Self-BLEU score than D2 and D3, indicating low redundancy and high diversity among the samples generated by D1.

\begin{table}[t]
    \centering
    \caption{Results corresponding to sequence-to-sequence models are shown. Model S1 is trained with sme-only using pretrained model weights on English language. S2 is trained from scratch with randomly initialized weights. S3 is trained with sme-only using pretrained model weights on Hungarian language. S4 is trained with sme-only using pretrained model weights on Finnish language. Overall, the training of sequence-to-sequence models for ULR languages appears less effective than decoder-only models. All models have used the Pegasus architecture.}
    \label{tab:sequence to sequence}
    \resizebox{.5\textwidth}{!}{%
    \begin{tabular}{@{}lccc@{}}
    \toprule
    Model & CE Loss  & Perplexity & Self-BLEU \\ \midrule
    S1    & 0.00015 & 1.00       & -         \\
    S2    & 7.01     & 1160       & 0.48      \\
    S3    & 0.16472     & 1.18       & -      \\
    S4    & 5.1374     & 170.28       & -      \\ \bottomrule
    \end{tabular}%
    }
\end{table}

In the second set of experiments, Table \ref{tab:sequence to sequence}, we tested the suitability of sequence-to-sequence based models for Sámi language modeling. We trained four different models: S1 was trained for sme-only over pretrained English language, S2 with random weights, S3 with Hungarian, S4 was pretrained with sme-only following pretraining for Finnish. It can be seen that pretraining with a more semantically close language (like Finnish) produces better results than training with randomized weights. In pretraining with multiple languages, it was found that English, which is semantically least closer to Sámi, produces even worse results than training with randomized weights. After this, Hungarian, which is closer to Sámi than English, produces improved results. On the same note, Finnish, which is even more semantically closer to Sámi than Hungarian, produces even better results. So, we can say that like decoder-only models, sequence-to-sequence-only models also benefit from multilingual training. However, if we compare S2 with D2, or S4 with D1 or D3, we could find that under both randomized and multilingual settings, decoder-only models perform better than sequence-to-sequence models.

In the following paragraphs, the generation results are presented. The results are formatted as Input and its translation in English followed by Completion and its translation in English. It can be seen that the text generated by decoder-only models is better than sequence-to-sequence. Moreover, pretraining over a solo language, D1, provides slightly more relevant generations over joint multilingual pretraining D2.

\subsection{Generation by SME only model, D1}
\noindent
\textit{Input-sme}: Beaivváš čuovgá álkit almmi ravddas.\\
\textit{Input-en}: The sun lightly shines from the edge of the sky.\\
\textit{Output-sme}: Beaivváš čuovgá álkit almmi ravddas. Lea dehálaš ahte sii geat leat riegádan ovdal, sáhttet hukset iežaset kultuvrra, luonddu ja birrasa, ja leat maid mielde ovddideamen oktavuođa. Dat buktá buoret vejolašvuođaid, muhto sáhttet veahkkin oaidnit maid mii vásihit.\\
\textit{Output-en}: The sun lightly shines from the edge of the sky. It is important that those born before can build their culture, nature and the environment, as well as contribute to the creation of one's bond. It brings better opportunities but can help to see what we're going through.\\

\noindent
\textit{Input-sme}: beatnagat oaguhit spáppaid párkkas.\\
\textit{Input-en}:  Dogs chasing balls in the park.\\
\textit{Output-sme}: beatnagat oaguhit spáppaid párkkas. Dat máksá ahte guolli galgá leat šaddan dakkár earuheapmen, mii muitala moivvi dahje suddan lea.\\
\textit{Output-en}: Dogs chasing balls in the park. It means that the fish must have become a distinction, which indicates the rot or melting.\\

\noindent
\textit{Input-sme}: Aroma dievvá gievkkana go mállása viežžá.\\
\textit{Input-en}:  The aroma fills up the kitchen when you take dinner.\\
\textit{Output-sme}: Aroma dievvá gievkkana go mállása viežžá.
Riddoguolástuslávdegoddi
sámi álbmoga
Sámedikki hálddahus
leat sierra čoahkkimat departemeanttain.
ja lea ráhkadan oktasaš plána\\
\textit{Output-en}: The aroma fills up the kitchen when you take dinner.
The Sametingets administration has separate meetings with the Ministry of the Sea.
and has drawn up a common plan 10.1\\

\noindent
\textit{Input-sme}: Láivvit biegga ruostu leavggaid čađa.\\
\textit{Input-en}:  A light breeze through the stairs of the stairs.\\
\textit{Output-sme}: Láivvit biegga ruostu leavggaid čađa. Goittotge, ahte maiddái Norgga beale sápmelaččat ožžot almmolaš doarjaga. Son lei vuosttažettiin mielde Norgga beale boazodoalu eanangeavaheami. Ja dát lea maiddái mielde maiddái eará sámi ássanhistorjjálaš orgána. Danin lean dán jagi hui somá go mun galgen álgit bargat boazodoaluin ja daid doallat ge eará eananosiid dan bokte go ođđa olbmot ledje leamaš mielde bargojoavkkus. Eai buot olbmot sáhttán šaddat doppe doppe ge.\\
\textit{Output-en}: A light breeze through the stairs of the stairs. Nevertheless, even the Sami in Norway are receiving public support. He was mainly involved in the use of the land in the Norwegian reindeer industry. And this is also included in other historical bodies of the Sami. Therefore, this year, I am very happy that I was going to be working on the reintroduction and also on the maintenance of other through the new members of the team. Not all men were able to grow there, either.

\subsection{Generation by multilingual model, D3}
\noindent
\textit{Input-sme}: Beaivváš čuovgá álkit almmi ravddas.\\
\textit{Input-en}:  The sun lightly shines from the edge of the sky.\\
\textit{Output-sme}: Beaivváš čuovgá álkit almmi ravddas. Su áhčči seahkalas, mii lei "Jiepmaluobmin" ja šattai "Jievju". Su áhčči ja losit galmmihanskábes lei juo seahkalas!
Sámediggi, Finnmárkku\\
\textit{Output-en}: The sun lightly shines from the edge of the sky. His father's coffin, which was called "Jiepmaluobmin" and became "Jievju". Her father and a heavier frying cape were already mixed!\\

\noindent
\textit{Input-sme}: beatnagat oaguhit spáppaid párkkas.\\
\textit{Input-en}:  Dogs chasing balls in the park.\\
\textit{Output-sme}: beatnagat oaguhit spáppaid párkkas.
leat buot golbma beali mat mearridit sin dárbbuid.
Sámi kulturmáhttu ja giellabirrasat galget leat oassin Sámi kulturfoanddas. Dat galget leat oassin dakkár guovddážat main lea buorre biras.\\
\textit{Output-en}: Dogs chasing balls in the park.
There are three factors that are determining their needs.
Saemien kultuvredaajroe jïh gïelebyjresh gelkieh meatan årrodh Saemien kultuvrefoentesne. They must be part of centres with a positive environment.\\

\noindent
\textit{Input-sme}: Aroma dievvá gievkkana go mállása viežžá.\\
\textit{Input-en}:  The aroma fills up the kitchen when you take dinner.\\
\textit{Output-sme}: Aroma dievvá gievkkana go mállása viežžá. Láigohanboađut leat gárvát ja dat galget merkejuvvot. Sihkkun lei maiddái dalle go son oaččui stuora borramuša. Dat eai gávdnon šat dálvi, dasgo daid sáhttá oastit. Son ii lean gal nu álki oastit, muhto fertii vuolgit dasa go son galggai oastit.\\
\textit{Output-en}: The aroma fills up the kitchen when you take dinner. The rental income is ready and must be registered. The delay also took place when she received a large amount of food. They didn't exist anymore this winter, because they could be purchased. He wasn't the easiest to buy, but he had to go for it when he was going to buy.\\

\noindent
\textit{Input-sme}: Láivvit biegga ruostu leavggaid čađa.\\
\textit{Input-en}:  A light breeze through the stairs of the stairs.\\
\textit{Output-sme}: Bargat dan ala vai árbediehtu dohkkehuvvo seamma dásis go dutkanvuđot máhttu vuođđun mearrádusaide luondduriggodagaid geavaheamis.
4.7.3.1.3 Doaibmadoarjja – njuolggodoarjagat
ja Davvi Romssa. Bargu álggahuvvo
Oza
3.7 Kultuvra
3.3.1.3 Dábálaš oahpponeavvut vuođđooahpahussii Tabealla 4.3.1.3 Oahpponeavvut sámegillii – jorgaleapmi ja heiveheapmi 4.3.1.4 Oahpponeavvut sámegielas – Ohcanvuđot doarjagat (1000 ru)
ja lea go dain leamaš vejolašvuohta
Sámi parlamentáralaš ráđđi\\
\textit{Output-en}: A light breeze through the stairs of the stairs.
Arbeide for equal recognition of traditional knowledge as based on research information as a basis for decisions on the use of natural resources.
4.7.3.1.3 Operational contributions - direct contributions and NorthernTroms. The project begins in Search 3.7 Culture 3.3.1.3 Ordinary learning materials for primary education Tabell 4.3.1.3 Learn materials in Finnish - translation and adaptation 4.3.1.4 Learn materials in Finnish - Applicable funding (1000 kr) and the availability of the programs.

\subsection{Generation by pretrained Pegasus model, S1}
\noindent
\textit{Input-sme}: Beaivváš čuovgá álkit almmi ravddas.\\
\textit{Input-en}:  The sun lightly shines from the edge of the sky.\\
\textit{Output-sme}: Beaivváš čuovgá álkit almmi ravddas.............................................."\\
\textit{Output-en}: The sun is lightly shining from the edge of the sky...."\\

\noindent
\textit{Input-sme}: beatnagat oaguhit spáppaid párkkas.\\
\textit{Input-en}:  Dogs chasing balls in the park.\\
\textit{Output-sme}: beatnagat oaguhit spáppaid párkkas.............................................."\\
\textit{Output-en}: Dogs chasing balls in the park ........"\\

\noindent
\textit{Input-sme}: Aroma dievvá gievkkana go mállása viežžá.\\
\textit{Input-en}:  The aroma fills up the kitchen when you take dinner.\\
\textit{Output-sme}: Aroma dievvá gievkkana go mállása viežžá.............................................."\\
\textit{Output-en}: Aroma fills the kitchen at dinner table...."\\

\noindent
\textit{Input-sme}: Láivvit biegga ruostu leavggaid čađa.\\
\textit{Input-en}:  A light breeze through the stairs of the stairs.\\
\textit{Output-sme}: Láivvit biegga ruostu leavggaid čađa.............................................."\\
\textit{Output-en}: "Lady wind through the flags of crust...."


\section{Discussion and Future Work}
This study revealed that training large language models for ultra low-resource languages like Sámi is a highly challenging process. Unlike low-resource languages, data generation is a highly complex process because for at least low-resource languages, some high-resource-pretrained LLMs like ChatGPT could be leveraged to meet the data deficiency; however, for ULRLs, the support from high-resource-pretrained LLMs is negligible and often not of high quality. Moreover, due to the very low number of speakers of ULRLs, manual data generation also becomes a very costly operation. These challenges are the main bottlenecks for language modeling in ULRLs. In this work, we compiled the online resources into a Sámi dataset for training and testing the performance of various classes of language models under different settings. We also tested the role of multilingual training since Sámi is very distinctly related to other families of languages. Due to the small size of the Sámi dataset, we conducted experiments on decoder-only LLMs and also on sequence-to-sequence models. The decoder-only experiments are described in Table \ref{tab:decoder only}. The rationale was to observe the performance in the presence of joint training vs. solo training to observe its sensitivity in the presence of other languages. The models were also tested for pretraining vs. training from scratch. The other set of experiments was carried out on sequence-to-sequence models as listed under Table \ref{tab:sequence to sequence}. Less overfitting was observed in decoder-only models compared to sequence-to-sequence models. Their performance and generations were better than sequence-to-sequence models, which indicates that they have better generalization abilities despite being bulkier architectures. It was also noticed that randomly initialized models are more capable of easily adapting to the ULR languages than pretrained models in highly semantically unrelated languages (see Table \ref{tab:sequence to sequence}). The generation results for sequence-to-sequence models were inferior to the decoder-only models. The generation results show that the decoder-only models were also not able to capture the context sometimes. However, legible text was being generated by the decoder-only models. It was also noticed that pretraining in sequence-to-sequence models was less effective than in decoder-only models. One possible reason for this may be the multi-headed flow of gradients and bulkier architectures. It was also noticed that pretraining with semantically distant languages (like English to Sámi) does more damage to model performance than semantically similar languages (like Finnish to Sámi). The training of LLMs on the Sámi dataset made us learn that the models need further finetuning with increased datasets. In our future work, we wish to explore the effects of bigger and contextually curated datasets for finetuning our pretrained base LLM models.

\subsection{Applications}
Sámi language models could be useful in a variety of potential applications across different fields and functionalities. This work is a small effort to foster a more inclusive and engaging digital experience for Sámi speakers. Below are a few possible applications that could be developed through the proposed work:

\begin{itemize}
    \item There is potential to develop accessibility tools such as speech-to-text and text-to-speech systems in the Sámi language.
    \item We are in the process of developing a robust plagiarism detector. This tool will be capable of analyzing Sámi texts to identify similarities and potential instances of plagiarism, proving invaluable in academic and professional settings.
    \item The models could be further customized to support essential NLP tasks like Named Entity Recognition (NER), which can help identify and categorize key information in Sámi texts into predefined categories.
    \item The models could be fine-tuned to enhance their ability to measure similarity between texts, which could be used in search functionalities, virtual assistants, providing more relevant and contextual responses.
    \item These models can be integrated into IoT devices to promote language-specific interactions.
    \item Models can be adapted for specific domains such as legal, medical, or historical contexts.    
\end{itemize}

Furthermore, to support ongoing research and application development, the dataset used for training our model, including tokens and other resources, has been made publicly available on our website (\href{https://bit.ly/samillm}{Project Page}). This enables researchers to access and utilize these resources for their research projects to contribute to the empowerment of the Sámi language community.

\section{Conclusion}
This work studies how recently developed large language models can be used to train for ultra low-resource languages like Sámi. We have explored the challenges involved in curating a specific dataset under a highly resource-constrained setting. It was learned that the existing LLMs claim to understand low-resource languages; however, their performance for ULRLs is very low and thus cannot be used for artificial data generation for training LLMs for ULRLs. We experimented with different training strategies and types of language modeling. It was found that pretraining with a semantically similar language positively contributes to the LLM training, while training with less semantically similar languages contributes negatively. We also found that training with one semantically related language is better than training with multiple less semantically related languages. Contrary to our belief, we also found that decoder-only models perform better than sequence-to-sequence models even when the training data is very limited. Guidelines for future work and possible applications have also been provided. We have released the data and code used for this study to further research.

\section{Acknowledgment}
We acknowledge the insightful linguistic discussions with Prof. Laura Janda at UiT. We are immensely grateful to Giellatekno \cite{giellatekno} and Sámi dieđalaš áigečála \cite{aigecala} for free Sámi resources.

\section{Funding}
This work was supported by the Research Council of Norway Project (nanoAI, Project ID: 325741), H2020 Project (OrganVision, Project ID: 964800).


\end{document}